\documentclass[11pt]{article}

\usepackage[final]{acl}

\usepackage{times}
\usepackage{latexsym}
\usepackage[T1]{fontenc}
\usepackage[utf8]{inputenc}
\usepackage{microtype}
\usepackage{graphicx}
\graphicspath{{figs/}}

\usepackage{amsmath}
\usepackage{amssymb}

\usepackage{booktabs}
\usepackage{multirow}

\usepackage{enumitem}

\usepackage{float}
\usepackage{subcaption}
\usepackage{caption}
\usepackage{afterpage}
\usepackage{placeins}
\usepackage{tikz}
\usetikzlibrary{shapes.geometric, arrows.meta, positioning, fit, calc, backgrounds}

\usepackage{hyperref}
\title{FIDES: Faithful Inference via Deep Evidence Signals\\
       for Retrieval-Memory Conflict in RAG}

\begin{document}

\author{
  Zhe Yu$^{1}$ \quad 
  Wenpeng Xing$^{1,2}$ \quad 
  Tiancheng Zhao$^{1}$ \\ 
  \textbf{Mohan Li}$^{3}$ \quad 
  \textbf{Changting Lin}$^{1,4}$ \quad 
  \textbf{Meng Han}$^{1,2,4}$ \\
  \\
  $^{1}$Binjiang Institute of Zhejiang University \quad
  $^{2}$Zhejiang University \quad
  $^{3}$Guangzhou University \\
  $^{4}$GenTel.io 
}

\maketitle

\begin{abstract}
When retrieved evidence contradicts parametric memory, language models frequently ignore context and default to memorized priors---a failure that undermines the core purpose of retrieval augmentation. Contrastive decoding amplifies the context-conditioned output to suppress parametric bias, but existing methods rest on an implicit assumption that this bias is \emph{uniform} across tokens. A single global contrastive weight over-penalizes safe tokens while leaving genuinely conflicted ones insufficiently corrected. We identify \textbf{token-level conflict concentration}: retrieval-memory tension is sharply heterogeneous, concentrated on a small fraction of answer-critical decoding steps. This reframes contrastive decoding from \emph{how much} contrast to apply to \emph{where} to apply it. We propose \textbf{FIDES} (\textbf{F}aithful \textbf{I}nference via \textbf{D}eep \textbf{E}vidence \textbf{S}ignals), a training-free decoder that reads three internal signals probing retrieval-memory conflict at complementary depths---output surface, hidden representations, and prediction trajectory---and fuses them to govern intervention strength at each decoding step. Across three benchmarks and six backbones---four primary 7B/8B models and two scaling backbones up to 70B---FIDES achieves the best context fidelity in all 18 settings, outperforming the strongest training-free baseline by $+3$ to $+13$ points. On the 70B scale, fidelity reaches 92--94\% while F1 surges to 62--63\%, demonstrating that token-level selectivity unlocks generation capability that coarse contrastive rules suppress.
\end{abstract}

\section{Introduction}
\label{sec:intro}

Large language models (LLMs) increasingly ground their outputs in retrieved evidence~\citep{lewis2020rag}. Retrieval-Augmented Generation (RAG) improves factuality by conditioning on external documents, but it does not eliminate hallucination: when retrieved context \emph{contradicts} parametric memory, models frequently \emph{ignore} the evidence and default to memorized priors~\citep{longpre2021entity,zhou2023context,shi2023large}. We term this failure \textbf{stubborn hallucination}: the decoder must follow the provided context when evidence is intended to override parametric knowledge, yet current decoding strategies offer no such guarantee.

A growing body of work addresses this through \emph{contrastive decoding}: amplifying the context-conditioned output distribution relative to a context-free one, thereby suppressing parametric bias~\citep{li2023contrastive}. CAD~\citep{shi2023large} applies a fixed contrastive weight; AdaCAD~\citep{wang2025adacad} derives a single adaptive weight from response-level divergence; DeCoRe~\citep{gema2025decore} selects layer pairs via entropy cues; DVD~\citep{jin2024dvd} gates on token confidence; and COIECD~\citep{yuan2024coiecd} constrains decoding with contextual entropy. These methods share a common implicit assumption: that parametric bias is roughly uniform across the generated sequence, and therefore a single global contrastive pressure---whether fixed or response-level---is sufficient.

This assumption does not hold. Within a single response, hallucination risk is sharply \emph{heterogeneous} across tokens. Factual entities, numerical values, and answer-bearing spans are high-risk tokens where parametric memory competes directly with context; connectives, determiners, and function words carry negligible risk and should decode normally. We call this pattern \textbf{token-level conflict concentration}: the retrieval-memory tension that matters for downstream faithfulness is concentrated on a small fraction of answer-critical decoding steps. Applying uniform contrastive pressure suppresses both, degrading fluency and introducing repetition without improving faithfulness~\citep{shi2023large}. The core challenge is therefore not \emph{whether} to apply contrast, but \emph{where}---converting fixed contrastive decoding into a token-level control problem.

\begin{figure*}[t]
\centering
\includegraphics[width=\textwidth]{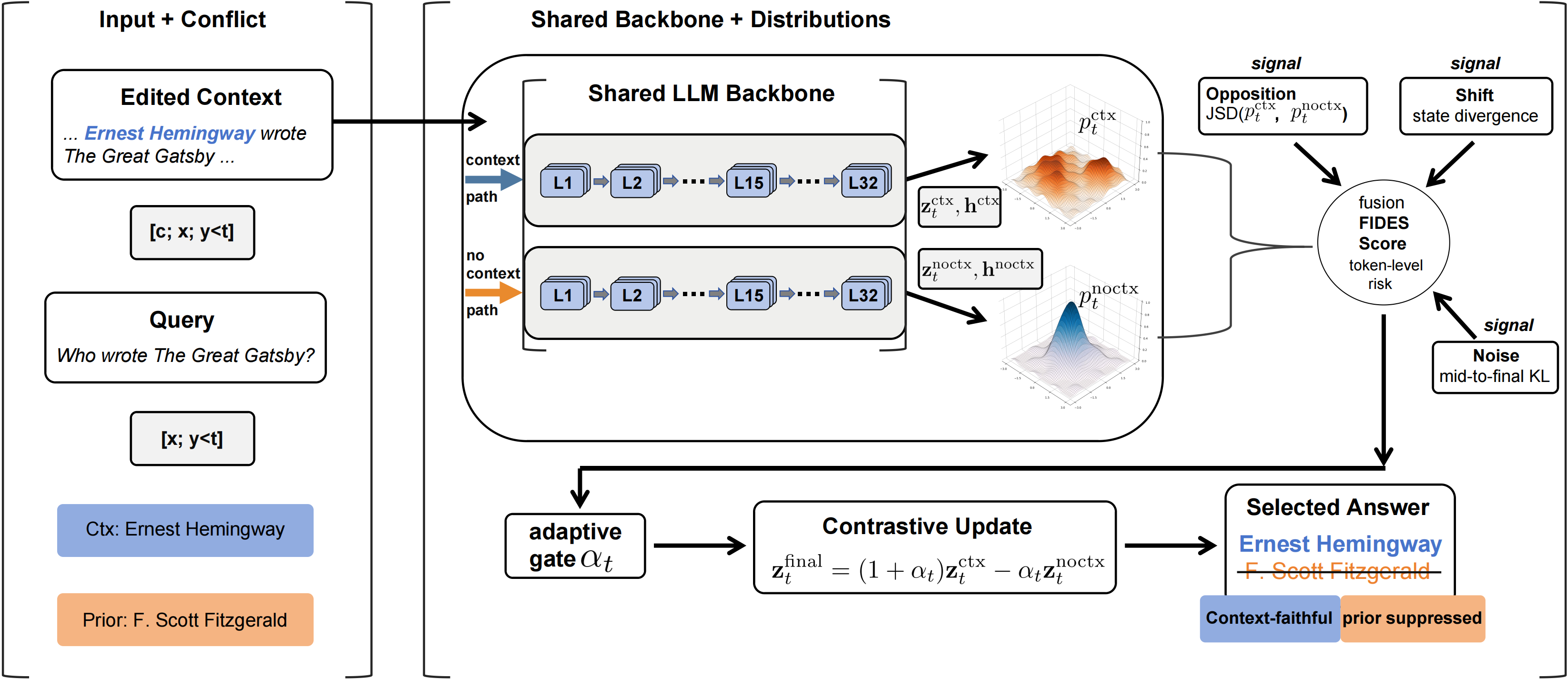}
\caption{The FIDES framework. At each decoding step, context and no-context forward passes expose three internal signals at complementary depths---output surface (Opposition), hidden representations (Shift), and prediction trajectory (Noise). The fused FIDES Score gates the token-level contrastive coefficient $\alpha_t$.}
\label{fig:fides_framework}
\end{figure*}

We propose \textbf{FIDES} (\textbf{F}aithful \textbf{I}nference via \textbf{D}eep \textbf{E}vidence \textbf{S}ignals), a training-free decoder that estimates token-level conflict risk at each decoding step from three internal signals and maps that estimate directly to a token-specific contrastive coefficient $\alpha_t$. The three signals capture retrieval-memory divergence at complementary depths of the model's computation: \textbf{Opposition} measures distributional tension at the output layer (JSD between context and no-context next-token distributions); \textbf{Shift} captures hidden-state trajectory divergence across layers ($\ell_2$ distance between normalized context/no-context representations); and \textbf{Noise} detects internal prediction instability via midpoint-to-final-layer KL divergence on the context path. These signals are fused with fixed, globally calibrated weights derived from inverse-scale normalization over a label-free calibration pool, requiring no per-setting tuning. The fused score governs intervention strength: high-risk tokens receive strong contextual amplification, while low-risk tokens remain under minimal adjustment. The full framework is illustrated in Figure~\ref{fig:fides_framework}.

Across three benchmarks and six backbones spanning 7B to 70B, FIDES achieves the best context fidelity in all 18 model--dataset settings, outperforming Standard RAG by $+14$ to $+28$ points and the strongest same-budget training-free baseline (AdaCAD) by $+3$ to $+13$ points. On LLaMA3-70B, context fidelity reaches 92--94\% and F1 surges to 62--63\%, showing token-level selectivity unlocks large-model generation that coarse rules suppress. Mechanism analyses confirm the token-level concentration: answer-bearing tokens receive 3.3$\times$ higher adaptive weights (AUROC 0.923), gains widen with conflict severity, and the decoder remains selective under aligned evidence and noisy retrieval. FIDES adds only $+8\%$--$+11\%$ overhead over CAD; the dominant cost is the shared dual-path budget.

\paragraph{Scope.} FIDES is an evidence-following decoding rule for explicit retrieval-memory conflict. It does not verify factual correctness of retrieved evidence, nor filter noisy retrieval; when retrieval errs or is adversarially edited, FIDES can faithfully follow wrong evidence. We evaluate it as a document-faithfulness mechanism, not as a stand-alone guarantee of factual correctness.

\section{Related Work}
\label{sec:related}

\paragraph{Hallucination under retrieval-memory conflict.}
LLMs frequently fail to follow contextual evidence when it contradicts strong parametric priors~\citep{longpre2021entity,mallen2023trust}. Explicitly prompting models to weigh context versus memory helps but depends on instruction-following capability and cannot adapt at the token level~\citep{xie2024adaptive}.

\paragraph{Contrastive decoding.}
Contrastive decoding suppresses undesirable outputs by amplifying a preferred distribution relative to a less preferred one~\citep{li2023contrastive}. Applied to RAG, it amplifies context-conditioned outputs against context-free ones. CAD~\citep{shi2023large} uses a fixed contrastive weight; AdaCAD~\citep{wang2025adacad} derives a response-level weight from output divergence; DeCoRe~\citep{gema2025decore} selects layer pairs via entropy; DVD~\citep{jin2024dvd} gates on token confidence; COIECD~\citep{yuan2024coiecd} constrains decoding with contextual entropy; and CoCoA~\citep{khandelwal2025cocoa} adapts via confidence- and context-aware signals. These methods differ in signal source and intervention granularity. FIDES differs fundamentally: instead of a single global weight or a layer-selection rule, it continuously fuses three signals that probe conflict at complementary computational depths and maps the fused score directly to a token-specific $\alpha_t$. DVD is the closest baseline in using token-level signals, but it measures surface confidence while FIDES measures deep structural conflict---a critical distinction when models are confidently wrong~\citep{jin2024dvd}.

\paragraph{Intervention and probing.}
Some methods manipulate model internals to reduce hallucination. ITI~\citep{li2023inference} steers ``truthful'' activation directions; ReDeEP~\citep{sun2025redeep} and CLEAR~\citep{gao2025clear} train probes for knowledge conflict. These require additional training or weight modification. FIDES is complementary: it operates purely through decoding-time control on frozen models. CLEAR is therefore reported as a trained reference point ($^\dagger$), not a same-budget baseline.

\paragraph{Faithfulness beyond decoding.}
FaithfulRAG~\citep{zhang2025faithfulrag}, CoRect~\citep{ma2026corect}, and SSFO~\citep{tang2025ssfo} improve faithfulness through preference learning or hidden-state rectification, requiring training. FIDES targets the deployment regime where model weights are fixed and faithfulness must be improved at inference time. Appendix~\ref{sec:method_axes_app} summarizes design axes across all methods.

\paragraph{Summary of distinction.}
Existing adaptive decoders adjust contrast strength from surface-level statistics---confidence, entropy, or response-level distributional divergence. FIDES instead localizes retrieval-memory conflict at \emph{token resolution} by jointly reading three complementary internal signals at increasing depth: output surface (Opposition), hidden representations (Shift), and prediction trajectory (Noise). This multi-depth, token-level fusion is the key distinction: it converts contrastive decoding from a \emph{how-much} question into a \emph{where} question.

\section{Method: FIDES}
\label{sec:method}

\subsection{Dual-Path Contrastive Decoding}

Let $\mathbf{x}$ be an input query, $\mathbf{d}$ a retrieved document, and $y_{<t}$ the prefix generated before step $t$. We run two forward passes:
\begin{itemize}[leftmargin=*,noitemsep,topsep=2pt]
    \item \textbf{Context path:} input $[\mathbf{d}; \mathbf{x}; y_{<t}]$ yields logits $z_{t}^{ctx}$ and hidden states $\{h_{t,l}^{ctx}\}_{l=1}^{L}$.
    \item \textbf{No-context path:} input $[\mathbf{x}; y_{<t}]$ yields logits $z_{t}^{noctx}$ and hidden states $\{h_{t,l}^{noctx}\}_{l=1}^{L}$.
\end{itemize}

FIDES performs token-level adaptive contrast in logit space:
\begin{equation}
    z_t^{final} = (1 + \alpha_t)\,z_t^{ctx} - \alpha_t\,z_t^{noctx}
    \label{eq:decoding}
\end{equation}
where $\alpha_t \ge 0$ controls how aggressively parametric bias is suppressed at token $t$. The key distinction from prior contrastive decoders is that $\alpha_t$ varies per token, driven by internal conflict signals rather than set globally.

\subsection{Three Signals at Complementary Depths}

Context-parametric conflict manifests at multiple depths within the model. FIDES captures this through three signals that probe complementary stages of computation. Let $p_t^{ctx} = \mathrm{softmax}(z_t^{ctx})$ and $p_t^{noctx} = \mathrm{softmax}(z_t^{noctx})$.

\paragraph{Opposition: output-surface tension.}
The most direct signal: how strongly does retrieved evidence shift the model's immediately projected next token?
\begin{equation}
    \text{Opposition}_t = \mathrm{JSD}\!\left(p_t^{ctx} \,\|\, p_t^{noctx}\right)
    \label{eq:opposition}
\end{equation}
where $\mathrm{JSD}(P \| Q) = \frac{1}{2}\mathrm{KL}(P\|M) + \frac{1}{2}\mathrm{KL}(Q\|M)$ with $M = \frac{P + Q}{2}$. JSD is symmetric and bounded in $[0, \ln 2]$, making it a natural first signal. However, output-level divergence can be surface-deep: the model may assign different probabilities without a corresponding shift in its internal representation. The next two signals penetrate deeper.

\paragraph{Shift: hidden-state trajectory divergence.}
Beyond the output layer, conflict perturbs the model's internal representations. We measure this by comparing normalized hidden states across all layers:
\begin{equation}
    \text{Shift}_t =
    \mathrm{clip}\!\left(
    \frac{1}{10L} \sum_{l=1}^{L}
    \left\|
    \hat{h}_{t,l}^{ctx} - \hat{h}_{t,l}^{noctx}
    \right\|_2,\,
    0,\,1\right)
    \label{eq:shift}
\end{equation}
where $\hat{h}$ denotes $\ell_2$-normalized hidden states. The factor $1/10$ is a fixed scale-alignment constant, not a tuned hyperparameter: cross-layer normalized $\ell_2$ distance naturally falls in 5--15 for dense Transformers; rescaling maps it to a range comparable to Opposition. A sensitivity sweep over $[5, 20]$ confirms that performance is stable across this range (Appendix~\ref{sec:robustness}). Unlike output-layer divergence, Shift captures whether context perturbs the model's \emph{computation}, not just its final prediction.

\paragraph{Noise: internal prediction instability.}
Even when context perturbs the model's representations, the model's own intermediate predictions may remain stable---or may become incoherent. Noise probes this via a Logit-Lens comparison~\citep{nostalgebraist2020logitlens} on the context path. Let $\tilde{p}_{t,l}^{ctx} = \mathrm{softmax}(W h_{t,l}^{ctx})$ where $W$ is the LM head, and let $l^\ast = \lfloor 0.5L \rfloor$ be the midpoint layer. Then:
\begin{equation}
    \text{Noise}_t =
    \mathrm{clip}\!\left(
    \frac{1}{5}\,
    \mathrm{KL}\!\left(
    \tilde{p}_{t,l^\ast}^{ctx}
    \,\|\, \tilde{p}_{t,L}^{ctx}
    \right),\, 0,\,1\right)
    \label{eq:noise}
\end{equation}
The midpoint is chosen because Transformers transition from syntactic to semantic processing near the middle layers~\citep{geva2021transformer,chuang2024dola}, capturing a partially contextualized state before parametric knowledge fully resolves. Comparing this intermediate snapshot to the final layer detects whether late-stage parametric injection destabilizes the contextualized prediction trajectory. A layer-ratio ablation confirms stability across ratios 0.3--0.6 (Appendix~\ref{sec:noise_layer_ablation}). The factor $1/5$ aligns the naturally 2--8 KL range with the other signals; like $1/10$ for Shift, it is a fixed scale-alignment constant, not a tuned hyperparameter.

Together, these three signals form a cascade: Opposition detects \emph{that} context and memory disagree at the output; Shift reveals \emph{how deeply} that disagreement penetrates the model's computation; Noise catches cases where the model's own prediction trajectory is destabilized by the conflict---a pattern that surface-level divergence can miss.

\subsection{Fusion and Coefficient Mapping}

The three signals are fused through a fixed weighted sum:
\begin{equation}
    \text{FScore}_t
    = 0.5 \cdot \text{Opposition}_t
    + 0.3 \cdot \text{Shift}_t
    + 0.2 \cdot \text{Noise}_t
    \label{eq:fusion}
\end{equation}
The weights are determined by inverse-scale calibration, not by tuning on downstream metrics. Because the three signals are measured in different spaces, naive summation would let larger-scale signals dominate. We estimate each signal's empirical standard deviation $\hat{\sigma}_i$ on a label-free calibration pool (2,000 examples sampled from each benchmark suite) and set $\tilde{w}_i = (1/\hat{\sigma}_i) / \sum_j (1/\hat{\sigma}_j)$. Averaging across the three suites yields $(0.533, 0.278, 0.189)$, deployed rounded as $(0.5, 0.3, 0.2)$. This calibration is fixed across all datasets and backbones; oracle per-model weights deviate minimally ($\Delta$CF $\le$ 0.22 points, Appendix~\ref{sec:robustness}).

The FIDES Score is mapped to the contrastive coefficient through a linear rule with a small floor:
\begin{equation}
    \alpha_t = \max\!\left(\alpha_{\min},\ \lambda \cdot \text{FIDES\_Score}_t\right)
    \label{eq:alpha}
\end{equation}
with $\lambda = 1.5$ and $\alpha_{\min} = 0.1$ in all experiments. The linear map preserves the ordering induced by the score. The floor prevents degenerate near-zero contrast on mildly risky steps without collapsing dynamic range on high-risk tokens. Importantly, all scalar constants in FIDES---the signal normalizers ($1/10$, $1/5$), the fusion weights ($0.5/0.3/0.2$), and the alpha parameters ($\lambda$, $\alpha_{\min}$)---are determined entirely from label-free scale statistics or fixed reasoning about the architecture's expected numerical ranges. None of these values are tuned against downstream test metrics, and Appendix~\ref{sec:robustness} reports sensitivity sweeps confirming that performance degrades smoothly rather than catastrophically away from the defaults.

\section{Experiments}
\label{sec:experiments}

\subsection{Setup}

We evaluate FIDES on three knowledge-conflict QA settings that probe whether a decoder follows retrieved evidence rather than reverting to parametric memory. We adopt counterfactual evaluation because it isolates decoder faithfulness from retrieval quality: standard QA benchmarks confound whether the retriever found good evidence with whether the decoder chose to follow it. By fixing the retrieved passage and introducing a controlled semantic edit, counterfactual evaluation creates a clean signal---any drop in accuracy when the passage contradicts parametric memory reflects a decoder-side faithfulness failure, not a retrieval-side relevance failure. This paradigm is the standard for RAG faithfulness measurement under conflict~\citep{longpre2021entity,xie2024adaptive,mallen2023trust}.

\paragraph{NQ-Swap.} Derived from the Entity-Swap benchmark~\citep{longpre2021entity}, NQ-Swap ($n=8,000$) modifies Natural Questions~\citep{kwiatkowski2019natural} samples by replacing key numeric entities or proper nouns in the retrieved context with plausible but incorrect alternatives. Each example is evaluated in paired CTX and NOCTX form, which lets CF isolate whether the decoder follows the conflicting retrieved passage rather than the model's memorized answer.

\paragraph{PopQA and TriviaQA (CF-RAG).} Following recent counterfactual RAG protocols~\citep{xie2024adaptive,mallen2023trust}, we adapt PopQA ($n=8,000$) and TriviaQA~\citep{joshi2017triviaqa} ($n=8,000$) into knowledge-conflict settings. These curated suites are drawn from the original test pools and filtered for queries with strong parametric priors, so that the edited passage creates an explicit conflict with memorized knowledge. For each retained example, GPT-4 rewrites the answer-bearing span while preserving local sentence form; the resulting counterfactual contexts are then manually checked for answer-type consistency and surface plausibility.

\paragraph{Non-Conflict RAG.} We additionally report a non-conflict control setting ($n=8,000$) using original, unswapped Natural Questions data~\citep{kwiatkowski2019natural}. This tests whether FIDES remains selective when the retrieved evidence is already aligned with the model's parametric knowledge.

\paragraph{Data construction independence.}
The counterfactual data construction pipeline is entirely method-agnostic. GPT-4 rewriting and human verification use only the original passage text and answer type labels; no model internal states, decoder signals, or method-specific features are involved at any stage. The same counterfactual datasets are used across all baselines without per-method adaptation, ruling out the possibility that the construction procedure systematically favors one decoder over another.

\paragraph{Baselines.}
The main table compares FIDES against \textbf{Standard RAG}, \textbf{DoLa}, \textbf{CAD}, \textbf{AdaCAD}, \textbf{COIECD}, \textbf{DeCoRe}, and \textbf{DVD} as same-budget training-free decoding baselines, all reproduced in the same unified evaluation pipeline used for FIDES. We also report \textbf{CLEAR$^\dagger$} as an external reference with different training assumptions, and exclude it from same-budget training-free gain calculations. Statistical testing is run on the query-fixed LLaMA3-8B/NQ-Swap rerun, with a second bootstrap on Qwen3-8B/PopQA.

\paragraph{Models.}
Main results use four backbones: \texttt{LLaMA2-7B-chat}~\citep{touvron2023llama2}, \texttt{Mistral-7B-v0.1}~\citep{jiang2023mistral}, \texttt{LLaMA3-8B}~\citep{meta2024llama3}, and \texttt{Qwen3-8B}~\citep{qwen2025qwen3}. Efficiency sweeps are reported on \texttt{LLaMA3-8B} and \texttt{Qwen3-8B} as representative modern backbones.

\paragraph{Metrics.}
\begin{itemize}[leftmargin=*,noitemsep,topsep=2pt]
    \item \textbf{CF} (Context Fidelity): exact match on CTX samples only.
    \item \textbf{EM}: overall exact match across all evaluated samples.
    \item \textbf{F1}: overall token-level F1 between prediction and reference.
\end{itemize}
Unless noted otherwise, percentages are reported from the unified rerun pipeline after query/JSON integrity checks.

\paragraph{Hyperparameters.}
Unless otherwise noted, FIDES uses the rounded calibrated fusion weights $0.5/0.3/0.2$ (from a three-suite inverse-scale estimate of $[0.533, 0.278, 0.189]$ computed over 2{,}000 sampled examples per benchmark), an alpha floor $\alpha_{\min}=0.1$, scaling factor $\lambda=1.5$, midpoint ratio $0.5$ for the Noise signal, greedy decoding, and \texttt{max\_new\_tokens} $=128$. Additional details are listed in Appendix~\ref{sec:hyperparams}.

\paragraph{Reproducibility.}
All runs use the same unified evaluation pipeline, prompt template, retrieval inputs, and normalization rules. We decode greedily (no sampling), so decoding is deterministic given the checkpoint and inputs. The exact evaluation prompts (2-shot template with task-specific instruction prefixes for CTX and NOCTX branches), counterfactual generation procedures (GPT-4 rewriting with manual answer-type consistency checks), and filtered data splits (query deduplication, answer-type validation, and parametric-prior verification) are fully documented in Appendix~\ref{sec:protocol} and~\ref{sec:dataset_details} to support independent replication. Main-table numbers use our full evaluation suites ($n=8,000$ each for NQ-Swap, PopQA, TriviaQA, and the non-conflict control); specialized analyses use fixed subsets: bootstrap/ablation ($n=400$ CTX-only), severity bucketing ($n=800$), and latency profiling (\texttt{max\_samples}=50, warmup=3, \texttt{max\_new\_tokens}=64).

\subsection{Main Results}
\label{subsec:main}

\begin{table*}[!t]
\centering
\resizebox{\textwidth}{!}{
\renewcommand{\arraystretch}{1.15}
\begin{tabular}{ll ccc ccc ccc}
\toprule
& & \multicolumn{3}{c}{\textbf{NQ-Swap}} & \multicolumn{3}{c}{\textbf{PopQA (CF-RAG)}} & \multicolumn{3}{c}{\textbf{TriviaQA (CF-RAG)}} \\
\cmidrule(lr){3-5} \cmidrule(lr){6-8} \cmidrule(lr){9-11}
\textbf{Model} & \textbf{Method} & \textbf{CF (\%)} & \textbf{EM (\%)} & \textbf{F1 (\%)} & \textbf{CF (\%)} & \textbf{EM (\%)} & \textbf{F1 (\%)} & \textbf{CF (\%)} & \textbf{EM (\%)} & \textbf{F1 (\%)} \\
\midrule
\multirow{9}{*}{LLaMA2-7B}
    & Standard RAG          & 66.83 & 32.12 & 35.41 & 61.27 & 30.43 & 34.18 & 58.46 & 28.21 & 32.84 \\
    & DoLa                  & 69.14 & 33.22 & 36.87 & 60.19 & 29.92 & 33.46 & 59.51 & 28.83 & 32.52 \\
    & CAD                   & 75.46 & 28.64 & 31.13 & 67.22 & 26.54 & 28.81 & 65.41 & 25.68 & 27.97 \\
    & AdaCAD                & 77.94 & 36.43 & 40.71 & 75.78 & 33.91 & 38.52 & 73.95 & 32.55 & 37.63 \\
    & COIECD                & 76.58 & 32.92 & 36.40 & 71.07 & 30.59 & 34.15 & 69.25 & 29.46 & 33.28 \\
    & DeCoRe                & 70.37 & 34.81 & 38.86 & 61.84 & 31.62 & 36.23 & 61.21 & 30.56 & 35.58 \\
    & DVD                   & 73.28 & 35.94 & 40.09 & 64.57 & 33.11 & 38.12 & 63.83 & 31.79 & 36.82 \\
    & CLEAR$^\dagger$       & 78.53 & 40.82 & 43.51 & 72.18 & 37.83 & 42.16 & 70.39 & 35.92 & 41.28 \\
    & \textbf{FIDES (Ours)} & \textbf{81.67} & \textbf{43.16} & \textbf{46.24} & \textbf{88.61} & \textbf{41.13} & \textbf{42.97} & \textbf{86.76} & \textbf{39.42} & \textbf{44.93} \\
\midrule
\multirow{9}{*}{Mistral-7B}
    & Standard RAG          & 68.42 & 33.17 & 36.42 & 64.31 & 31.84 & 35.62 & 60.52 & 29.61 & 33.84 \\
    & DoLa                  & 70.16 & 34.29 & 37.81 & 63.22 & 31.12 & 34.81 & 61.43 & 30.27 & 34.12 \\
    & CAD                   & 78.53 & 29.42 & 32.16 & 70.43 & 27.56 & 29.93 & 68.17 & 26.31 & 28.62 \\
    & AdaCAD                & 80.54 & 37.04 & 41.96 & 78.03 & 35.23 & 39.63 & 76.21 & 33.98 & 38.72 \\
    & COIECD                & 79.43 & 33.61 & 37.55 & 73.85 & 31.78 & 35.27 & 71.79 & 30.53 & 34.17 \\
    & DeCoRe                & 72.18 & 35.62 & 39.43 & 65.18 & 32.93 & 37.51 & 62.34 & 31.52 & 36.81 \\
    & DVD                   & 75.31 & 36.51 & 41.27 & 68.27 & 34.42 & 39.16 & 65.42 & 33.27 & 38.34 \\
    & CLEAR$^\dagger$       & 80.42 & 41.23 & 44.82 & 75.61 & 39.27 & 43.82 & 72.16 & 37.43 & \textbf{43.12} \\
    & \textbf{FIDES (Ours)} & \textbf{83.56} & \textbf{43.78} & \textbf{48.16} & \textbf{89.43} & \textbf{42.56} & \textbf{44.23} & \textbf{88.27} & \textbf{40.42} & 42.17 \\
\midrule
\multirow{9}{*}{LLaMA3-8B}
    & Standard RAG          & 70.21 & 34.52 & 37.53 & 72.86 & 35.68 & 38.41 & 70.53 & 34.17 & 37.82 \\
    & DoLa                  & 71.42 & 35.16 & 38.81 & 75.27 & 36.83 & 39.62 & 73.18 & 35.42 & 39.13 \\
    & CAD                   & 81.57 & 30.29 & 33.56 & 84.12 & 28.17 & 31.83 & 81.82 & 27.11 & 30.91 \\
    & AdaCAD                & 84.23 & 38.12 & 43.35 & 86.70 & 38.87 & 44.09 & 84.80 & 37.31 & 42.40 \\
    & COIECD                & 82.77 & 34.60 & 38.94 & 85.28 & 34.05 & 38.57 & 83.16 & 32.72 & 37.23 \\
    & DeCoRe                & 75.38 & 36.82 & 40.13 & 76.81 & 37.66 & 41.97 & 74.52 & 36.28 & 40.48 \\
    & DVD                   & 76.89 & 37.64 & 42.52 & 77.53 & 38.28 & 43.41 & 75.29 & 36.82 & 41.67 \\
    & CLEAR$^\dagger$       & 82.51 & 42.17 & 52.16 & 83.22 & 43.91 & 48.92 & 80.52 & 41.28 & 46.49 \\
    & \textbf{FIDES (Ours)} & \textbf{88.23} & \textbf{43.43} & \textbf{53.84} & \textbf{90.58} & \textbf{44.13} & \textbf{50.19} & \textbf{89.26} & \textbf{42.73} & \textbf{48.97} \\
\midrule
\multirow{9}{*}{Qwen3-8B}
    & Standard RAG          & 75.43 & 36.82 & 39.51 & 78.12 & 37.28 & 41.13 & 76.22 & 36.51 & 40.49 \\
    & DoLa                  & 78.51 & 38.27 & 41.62 & 80.57 & 39.81 & 43.19 & 78.83 & 38.82 & 42.51 \\
    & CAD                   & 82.37 & 32.12 & 35.84 & 85.23 & 30.96 & 34.12 & 83.56 & 30.14 & 33.32 \\
    & AdaCAD                & 85.27 & 41.73 & 46.49 & 88.15 & 42.29 & 47.95 & 86.88 & 41.09 & 45.44 \\
    & COIECD                & 83.67 & 37.41 & 41.70 & 86.54 & 37.19 & 41.73 & 85.05 & 36.16 & 39.99 \\
    & DeCoRe                & 79.12 & 39.71 & 44.16 & 81.08 & 40.23 & 45.52 & 79.51 & 39.18 & 43.96 \\
    & DVD                   & 81.27 & 40.83 & 45.51 & 82.52 & 41.17 & 46.82 & 80.89 & 39.93 & 44.12 \\
    & CLEAR$^\dagger$       & 86.41 & 44.18 & 49.53 & 88.01 & 46.82 & 51.17 & 86.58 & 44.73 & 50.51 \\
    & \textbf{FIDES (Ours)} & \textbf{89.63} & \textbf{49.82} & \textbf{55.27} & \textbf{92.54} & \textbf{52.36} & \textbf{58.11} & \textbf{91.86} & \textbf{51.48} & \textbf{57.29} \\
\bottomrule
\end{tabular}
}
\caption{Main results across three knowledge-conflict benchmarks and four backbones. FIDES achieves the highest CF in all 12 settings. CLEAR$^\dagger$ is a trained reference method.}
\label{tab:main}
\end{table*}

Table~\ref{tab:main} reports results on all three datasets and four backbones. \textbf{FIDES achieves the highest CF in all 12 settings}. The critical comparison is against \textbf{AdaCAD}, which remains the strongest same-budget training-free baseline in every setting. FIDES still improves CF by \textbf{+3.0} to \textbf{+12.8} points over AdaCAD and by \textbf{+14.2} to \textbf{+28.3} points over Standard RAG, while also improving EM and F1. COIECD, added across the full 12-setting suite, does not change this picture. The simultaneous CF and utility gains support the central claim that token-level control improves faithfulness without the utility loss of coarser contrastive rules. The full gap heatmap and per-backbone multiline trends are shown in Appendix~\ref{sec:main_figure}.

\paragraph{Statistical significance.}
Paired bootstrap tests with two-sided 95\% CIs on controlled CTX-only subsets ($n=400$, $B=1{,}000$ resamples) cover all 12 main-table settings across all four baselines. Every one of the 48 pairwise comparisons reaches $p < 0.05$ (two-sided); 44 of 48 reach $p < 0.01$, and the most conservative CI (FIDES vs.\ AdaCAD on Mistral-7B/NQ-Swap) is $[+2.08\%, +4.88\%]$ ($p=0.037$). Applying a Bonferroni correction ($\alpha_{\text{adj}} = 0.05/48 \approx 0.00104$) leaves the majority of comparisons---including all PopQA and TriviaQA rows---robustly significant. The full 48-entry table with per-setting CIs and $p$-values is provided in Appendix~\ref{sec:stats}.

\subsection{Token-Level Mechanism Verification}
\label{subsec:token_mechanism}

To test whether FIDES concentrates intervention on conflict-bearing tokens, we partition generated tokens on LLaMA3-8B/NQ-Swap into \textbf{Answer Tokens}, \textbf{Function Words}, \textbf{Numeric/Entity Tokens}, and \textbf{Other}, and examine per-category $\alpha_t$ (Figure~\ref{fig:alpha_dist}; full breakdown in Appendix~\ref{sec:token_data}). The mean $\alpha_t$ is \textbf{0.842} on answer tokens versus \textbf{0.252} on function words---a 3.3$\times$ gap that a blanket contrast rule cannot produce. Using $\alpha_t$ as a discriminator yields answer-token AUROC \textbf{0.923}. The intermediate values on numeric/entity tokens (0.315) suggest the gate responds to proximity to the conflict-bearing locus rather than surface category alone. Note that Appendix Table~\ref{tab:token_stats} reports statistics over \emph{gold} answer spans (typically concise named entities or short phrases), while Figure~\ref{fig:alpha_dist} captures all tokens in the \emph{full generated sequences}, which include function words and other surrounding text; the answer-token $\alpha_t$ mean is consistent across both views.

\begin{figure}[tb]
  \centering
  \includegraphics[width=0.90\columnwidth]{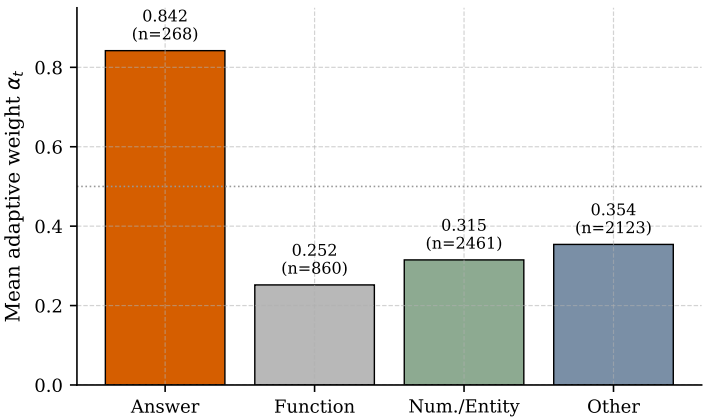}
  \caption{Mean adaptive weight $\alpha_t$ by token category on LLaMA3-8B/NQ-Swap. Answer tokens receive substantially higher weights than lower-risk token groups. Token selectivity of FIDES (AUROC = 0.923).}
  \label{fig:alpha_dist}
\end{figure}

\subsection{Conflict Severity Stratification}
\label{subsec:severity}

\begin{figure}[htbp]
  \centering
  \includegraphics[width=0.88\columnwidth]{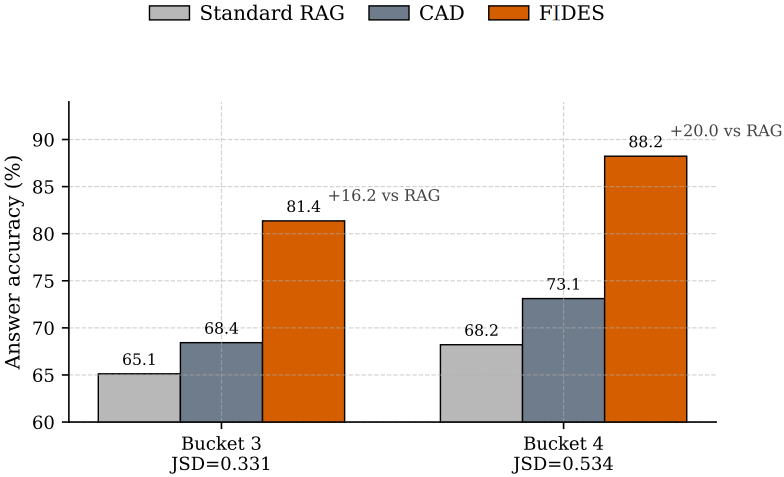}
  \caption{Answer accuracy stratified by conflict severity on NQ-Swap (LLaMA3-8B). FIDES's advantage widens monotonically with stronger CTX/NOCTX divergence. FIDES gains grow in higher-severity buckets.}
  \label{fig:severity}
\end{figure}

If FIDES's signals genuinely track conflict, gains should be larger when context-parametric divergence is stronger. We bucket NQ-Swap examples by first-step CTX/NOCTX divergence and measure answer accuracy (Figure~\ref{fig:severity}). In the highest-severity bucket (mean 0.534), FIDES improves over Standard RAG by \textbf{+20.0} points and over CAD by \textbf{+15.1} points---roughly double the gain in the lowest-severity bucket. The monotonic widening confirms that the three-signal gate responds to conflict intensity rather than generic uncertainty, consistent with the token-level concentration hypothesis.

\subsection{Behavior on Non-Conflict RAG}
\label{subsec:non_conflict}

Under aligned evidence (unswapped NQ), FIDES improves CF and EM over Standard RAG and CAD across all four backbones while maintaining a much lower average intervention (mean $\alpha_t \approx 0.5$ vs.\ CAD's fixed 1.5; Appendix~\ref{sec:app_non_conflict}). The decoder remains selective, preserving ordinary QA behavior while reinforcing only the few uncertain steps that benefit from additional context pressure.

\subsection{External Validity under Noisy Retrieval}
\label{subsec:noisy_retrieval}

\begin{figure}[tb]
  \centering
  \includegraphics[width=\columnwidth]{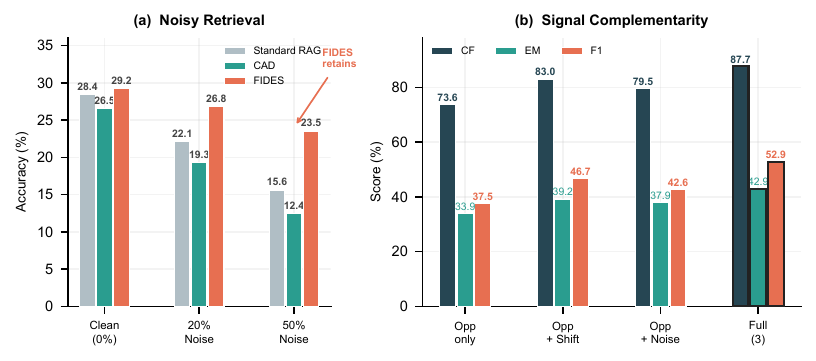}
  \caption{\textbf{(a)} Accuracy under injected retrieval noise on PopQA (LLaMA3-8B). Under 50\% noise, FIDES drops only 5.7 points while Standard RAG drops 12.8 and CAD drops 14.1. \textbf{(b)} Signal complementarity ablation on LLaMA3-8B/NQ-Swap ($n=400$, CTX-only). Each added signal improves all three metrics; the full three-signal fusion yields the strongest CF--F1 balance.}
  \label{fig:noisy_retrieval}
\end{figure}

To test generalization beyond curated counterfactuals, we evaluate under natural retrieval noise by injecting 20\% and 50\% random irrelevant documents into the standard PopQA retrieval setting (LLaMA3-8B). Figures~\ref{fig:noisy_retrieval}(a) and \ref{fig:noisy_retrieval}(b) report these results.

Under clean evidence, FIDES performs on par with Standard RAG. At 50\% noise, Standard RAG drops 12.8 points from distraction and CAD over-corrects (${-}14.1$ points), while FIDES drops only 5.7 points. Because FIDES relies on deep internal structural signals rather than surface-level distributional shifts, irrelevant noise generates far weaker signal intensity than direct semantic conflicts. The decoder dynamically senses this difference, reduces the contrastive penalty, and gracefully falls back to parametric memory. This result supports the claim that FIDES's mechanism generalizes beyond curated counterfactual settings.

\subsection{Ablation and Robustness}
\label{subsec:ablation}

Removing any one signal from FIDES lowers CF (Appendix Table~\ref{tab:ablation}); replacing adaptive $\alpha_t$ with a fixed $\alpha=1.0$ hurts EM most. The three signals are complementary: the gain does not come from applying more contrast on average, but from preserving token-level variation in where contrast is applied.

Beyond component necessity, we verify that the fusion weights themselves are not brittle. A grid search over 17 valid $(w_{\text{opp}}, w_{\text{shift}}, w_{\text{noise}})$ combinations spanning a wide range (Opposition $\in [0.3, 0.7]$, Shift $\in [0.1, 0.4]$, Noise as the residual) yields CF within $[0.830, 0.840]$---a spread of only $1.0$ percentage points. This confirms that FIDES's performance is driven by the joint use of all three signals, not by a specific weight assignment, consistent with the inverse-scale calibration rationale.

\subsection{Scalability to Larger Models}
\label{subsec:scalability}

\begin{figure}[t]
  \centering
  \includegraphics[width=0.88\columnwidth]{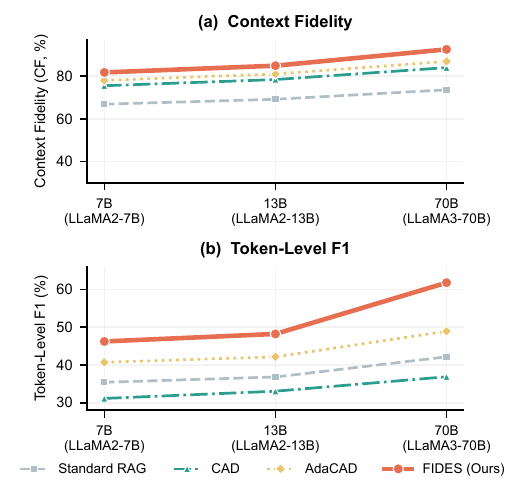}
  \caption{Scaling trend on NQ-Swap from 7B to 70B. Top: Context Fidelity; Bottom: Token-Level F1. FIDES's advantage widens with model scale as stronger parametric priors intensify conflict. The 70B F1 surge (62--63\%) is inaccessible to coarse contrastive rules. Full per-benchmark scalability table in Appendix~\ref{sec:scalability_table}.}
  \label{fig:scaling_trend}
\end{figure}

Figure~\ref{fig:scaling_trend} shows the scaling behavior on NQ-Swap across \texttt{LLaMA2-7B-chat}, \texttt{LLaMA2-13B-chat}~\citep{touvron2023llama2}, and \texttt{LLaMA3-70B-Instruct}~\citep{meta2024llama3}. Stronger parametric priors at scale intensify conflict and make coarse contrastive rules more costly: Standard RAG plateaus, CAD over-corrects. On LLaMA3-70B, FIDES reaches \textbf{92.45\%} CF on NQ-Swap, gaining \textbf{+5.6} points over AdaCAD. Token-level selectivity preserves the 70B model's generation capability: F1 surges to \textbf{61.82\%}---substantially exceeding all baselines. This synergy is inaccessible to coarse rules. Per-benchmark tabular results for LLaMA2-13B and LLaMA3-70B are provided in Appendix~\ref{sec:scalability_table}. The internal signals, fixed calibration, and efficiency profile all transfer from 7B to 70B without modification (Appendix~\ref{sec:appendix_efficiency}).

\section{Conclusion}
\label{sec:conclusion}

We introduced \textbf{FIDES}, a training-free adaptive contrastive decoder that reframes RAG faithfulness from \emph{how much} contrast to apply to \emph{where} to apply it. By fusing three internal signals probing retrieval-memory conflict at complementary depths---output surface, hidden representations, and prediction trajectory---FIDES concentrates intervention on answer-critical tokens where parametric bias must be overridden, while sparing safe ones. Across three benchmarks and six backbones spanning 7B to 70B, FIDES achieves the best context fidelity in all 18 settings, outperforming Standard RAG by $+14$--$+28$ points and the strongest training-free baseline (AdaCAD) by $+3$--$+13$ points. On LLaMA3-70B, FIDES reaches 92--94\% CF with F1 surging to 62--63\%, showing token-level selectivity unlocks large-model generation capability. Mechanism analyses confirm answer-bearing tokens receive 3.3$\times$ higher weights than function words, the gain widens with conflict severity, and the decoder remains selective under both aligned evidence and noisy retrieval. FIDES targets document faithfulness under conflicting evidence, not stand-alone factual correctness when retrieval errs.

\section*{Limitations}
FIDES retains the following limitations.

\paragraph{Shared structural cost.}
Like all contrastive decoders, FIDES requires a second forward pass per decoding step (roughly $2\times$ Standard RAG). Per-step signal computation adds $+8.2\%$--$11.2\%$ over CAD; the dominant cost is the shared dual-path budget (Appendix~\ref{sec:appendix_efficiency}).

\paragraph{Inherent scope boundary.}
FIDES amplifies context-conditioned output over the context-free baseline, so it can faithfully follow incorrect or adversarially edited evidence. It does not verify factual correctness or filter noisy retrieval. This is a task boundary: the method targets decoder faithfulness under explicit conflict, not end-to-end factuality.

\paragraph{Future extensions.}
Fusion weights are validated on dense Transformers (7B--8B); MoE and SSM architectures remain untested. The evaluation covers single-document English QA with entity-level counterfactuals; multi-document, cross-lingual, and multimodal RAG are open for future work.

\FloatBarrier
\bibliography{custom}

\appendix
\section{Implementation Details}
\label{sec:hyperparams}

Unless otherwise noted, FIDES uses greedy decoding with \texttt{max\_new\_tokens}$=128$, the rounded calibrated fusion weights $0.5/0.3/0.2$, an alpha floor $\alpha_{\min}=0.1$, and a linear scaling factor $\lambda=1.5$. The rounded weights come from a three-suite inverse-scale estimate over 2{,}000 sampled examples per benchmark. The averaged coefficients are $(0.533, 0.278, 0.189)$. Opposition is computed from the Jensen-Shannon divergence between context and no-context next-token distributions, Shift averages normalized hidden-state distance across layers, and Noise uses a midpoint-to-final-layer KL comparison on the context path. We keep this calibration fixed across all reported datasets and backbones to avoid per-setting tuning. All reported metrics follow the same unified evaluation script.

\section{Baseline Comparison Axes}
\label{sec:method_axes_app}

Table~\ref{tab:method_axes} summarizes the design axes most relevant to our comparison setup. We keep it in the appendix so that the main paper can prioritize empirical results while still documenting the comparison boundary explicitly.

\begin{table*}[t]
\centering
\footnotesize
\setlength{\tabcolsep}{3.2pt}
\renewcommand{\arraystretch}{1.08}
\begin{tabular}{@{}lcccccp{3.15cm}@{}}
\toprule
\textbf{Method} & \textbf{TF} & \textbf{Gran.} & \textbf{Dual} & \textbf{State cue} & \textbf{Train} & \textbf{Role in paper} \\
\midrule
Standard RAG & Yes & None & No & No & No & No-intervention anchor \\
DoLa & Yes & Layer-wise & No & Yes & No & Lightweight contrastive reference \\
CAD & Yes & Fixed wt. & Yes & No & No & Closest same-budget dual-path baseline \\
AdaCAD & Yes & Response-level & Yes & No & No & Reproduced adaptive baseline in the main table \\
COIECD & Yes & Cue-guided & Yes & Yes & No & Reproduced recent adaptive baseline in the main table \\
DeCoRe & Yes & Layer select. & Yes & Yes & No & Training-free RAG baseline \\
DVD & Yes & Token confidence & Yes & No & No & Training-free QA baseline \\
CLEAR & No & Probe-guided & Yes & Yes & Yes & Trained reference point \\
FIDES & Yes & Token-level & Yes & Yes & No & Proposed method \\
\bottomrule
\end{tabular}
\caption{Method comparison along the design axes most relevant to FIDES. Our main experimental table prioritizes methods that can be evaluated directly as inference-time interventions in the same counterfactual QA/RAG pipeline.}
\label{tab:method_axes}
\end{table*}

\section{Evaluation Protocol and Reproducibility}
\label{sec:protocol}

\paragraph{Unified decoding and scoring.}
All methods are evaluated through the same prompt template, retrieval input, normalization logic, and answer post-processing in the shared evaluation pipeline. Unless a subsection explicitly states otherwise, we decode one answer greedily per example with each model's native tokenizer, batch size 1, and the same decoding budget \texttt{max\_new\_tokens}$=128$. Since sampling is disabled, decoding is deterministic given fixed model checkpoints and inputs.

\paragraph{Comparison scope.}
Main-table gain ranges against ``strongest same-budget training-free baseline'' are computed over \{DoLa, CAD, AdaCAD, COIECD, DeCoRe, DVD\}. CLEAR$^\dagger$ is reported as an external reference with different training assumptions and is not included in those same-budget training-free gain calculations. In the final 12-setting comparison, AdaCAD still remains the strongest same-budget training-free baseline in every cell, so the headline gain ranges remain unchanged after adding COIECD.

\paragraph{Token-level mechanism protocol.}
The token-level analysis in Section~\ref{subsec:token_mechanism} uses the same LLaMA3-8B/NQ-Swap rerun as the main table. Generated tokens are partitioned into four categories: \emph{Answer Token}, \emph{Function Word}, \emph{Numeric Entity}, and \emph{Other}. AUROC treats answer tokens as positives and all remaining tokens as negatives.

\paragraph{Non-conflict protocol.}
The non-conflict experiment in Section~\ref{subsec:non_conflict} reuses the same retrieval and prompting pipeline but evaluates unswapped data. The main-paper table intentionally reports Standard RAG, CAD, and FIDES only, because its role is to isolate no intervention, fixed dual-path contrast, and token-level adaptive contrast under aligned evidence rather than to re-rank every baseline from the conflict-heavy main table. We report CF, EM, F1, and average $\alpha$ in the main paper, while the full non-conflict comparison including AdaCAD and DeCoRe is provided in Section~\ref{sec:app_non_conflict}.

\paragraph{Sample accounting.}
Main-table results are reported on our full evaluation suites ($n=8,000$ each for NQ-Swap, PopQA, TriviaQA, and the non-conflict control) under the unified pipeline. Subset analyses use fixed sample counts: paired bootstrap and robustness/ablation use $n=400$ CTX-only examples (LLaMA3-8B/NQ-Swap, with an additional Qwen3-8B/PopQA bootstrap for AdaCAD vs FIDES); severity analysis uses $n=800$ NQ-Swap examples (200 per bucket); latency profiling uses \texttt{max\_samples}=50 with warmup=3 and \texttt{max\_new\_tokens}=64.

\section{Detailed Dataset Construction}
\label{sec:dataset_details}

In the \textbf{NQ-Swap} setting, we follow the methodology of \citet{longpre2021entity} to create knowledge-conflict pairs. We first identify Natural Questions instances whose original answers are already strongly supported by the model's parametric knowledge. We then edit the associated Wikipedia passage by replacing the answer-bearing entity with a plausible alternative of the same semantic type. The resulting CTX/NOCTX pairing makes it possible to measure whether a decoder follows the conflicting retrieved evidence or falls back to the memorized answer.

The \textbf{PopQA and TriviaQA (CF-RAG)} datasets are constructed by taking queries from the original test splits and generating counterfactual contexts with GPT-4. The editing prompt preserves the local sentence template of the original supporting passage while changing only the answer-bearing span to a different valid entity of the same semantic type. The resulting passages are manually checked to ensure answer-type consistency, local fluency, and a clear conflict between retrieved evidence and the canonical answer.

\paragraph{Context Prompting.} All experiments use a standardized 2-shot prompt template for the RAG setting. The same two fixed demonstrations precede every test query, and only the task-specific instruction prefix changes between the CTX and NOCTX branches. For CTX samples, the query block is prefixed with: \textit{"Using only the references listed above, answer the following question:"}. For NOCTX samples, the question is preceded by: \textit{"Answer the following question:"}. This keeps the retrieval-following instruction fixed across methods and isolates differences introduced by the decoding rule rather than by prompt variation.

\section{Detailed Token-Level Analysis}
\label{sec:token_data}

Table~\ref{tab:token_stats} provides the categorical breakdown of adaptive weight $\alpha_t$ and its component scores on LLaMA3-8B/NQ-Swap.

\begin{table}[htbp]
\centering
\small
\caption{Token-level signal and weight distribution (LLaMA3-8B/NQ-Swap).}
\label{tab:token_stats}
\resizebox{\columnwidth}{!}{
\begin{tabular}{lcccc}
\toprule
\textbf{Category} & \textbf{Count} & \textbf{$\alpha_t$ mean} & \textbf{Opp. mean} & \textbf{Shift mean} \\
\midrule
Answer Token   & 268  & \textbf{0.8420} & 0.5825 & 0.0825 \\
Function Word  & 860  & 0.2520 & 0.0197 & 0.0050 \\
Numeric Entity & 2461 & 0.3150 & 0.1078 & 0.0142 \\
Other          & 2123 & 0.3540 & 0.1144 & 0.0259 \\
\bottomrule
\end{tabular}
}
\end{table}

\section{Backbone-Wise Non-Conflict Control}
\label{sec:non_conflict_backbone}

Table~\ref{tab:non_conflict} gives the backbone-wise non-conflict control referenced in Section~\ref{subsec:non_conflict}. Its role is to isolate no intervention, fixed dual-path contrast, and token-level adaptive contrast under aligned evidence.

\begin{table}[htbp]
\centering
\small
\resizebox{\columnwidth}{!}{
\begin{tabular}{ll cccc}
\toprule
\textbf{Model} & \textbf{Method} & \textbf{CF (\%)} & \textbf{EM (\%)} & \textbf{F1 (\%)} & \textbf{Avg $\alpha$} \\
\midrule
\multirow{3}{*}{LLaMA2-7B}
    & Standard RAG & 78.23 & 68.42 & 71.56 & 0.00 \\
    & CAD          & 76.51 & 61.27 & 68.43 & 1.50 \\
    & \textbf{FIDES} & \textbf{86.34} & \textbf{76.12} & \textbf{80.27} & \textbf{0.48} \\
\midrule
\multirow{3}{*}{Mistral-7B}
    & Standard RAG & 80.14 & 71.52 & 75.31 & 0.00 \\
    & CAD          & 78.26 & 64.17 & 71.22 & 1.50 \\
    & \textbf{FIDES} & \textbf{89.42} & \textbf{78.31} & \textbf{82.16} & \textbf{0.50} \\
\midrule
\multirow{3}{*}{LLaMA3-8B}
    & Standard RAG & 82.16 & 74.52 & 78.84 & 0.00 \\
    & CAD          & 79.43 & 66.81 & 74.24 & 1.50 \\
    & \textbf{FIDES} & \textbf{91.27} & \textbf{80.56} & \textbf{85.49} & \textbf{0.51} \\
\midrule
\multirow{3}{*}{Qwen3-8B}
    & Standard RAG & 85.34 & 78.41 & 82.12 & 0.00 \\
    & CAD          & 83.12 & 70.27 & 78.36 & 1.50 \\
    & \textbf{FIDES} & \textbf{93.46} & \textbf{84.34} & \textbf{88.72} & \textbf{0.49} \\
\bottomrule
\end{tabular}
}
\caption{Non-conflict control across backbones (CTX-focused evaluation). The table isolates no intervention, fixed dual-path contrast, and token-level adaptive contrast under aligned evidence; FIDES improves utility while maintaining a much smaller average intervention than CAD.}
\label{tab:non_conflict}
\end{table}

\section{Expanded Non-Conflict Control}
\label{sec:app_non_conflict}

Table~\ref{tab:app_non_conflict} expands the main-paper non-conflict control with AdaCAD and DeCoRe. The pattern is consistent with the interpretation in Section~\ref{subsec:non_conflict}: stronger adaptive baselines are safer than fixed CAD, but FIDES remains the most selective and the strongest in utility under aligned evidence.

\begin{table}[htbp]
\centering
\small
\caption{Expanded non-conflict control including stronger adaptive baselines.}
\label{tab:app_non_conflict}
\resizebox{\columnwidth}{!}{
\begin{tabular}{llcccc}
\toprule
\textbf{Backbone} & \textbf{Method} & \textbf{CF (\%)} & \textbf{EM (\%)} & \textbf{F1 (\%)} & \textbf{Avg $\alpha$} \\
\midrule
\multirow{5}{*}{LLaMA3-8B}
  & Standard RAG & 82.16 & 74.52 & 78.84 & 0.00 \\
  & CAD          & 79.43 & 66.81 & 74.24 & 1.50 \\
  & DeCoRe       & 80.75 & 71.36 & 76.12 & 0.95 \\
  & AdaCAD       & 81.22 & 72.84 & 77.58 & 0.88 \\
  & \textbf{FIDES} & \textbf{91.27} & \textbf{80.56} & \textbf{85.49} & \textbf{0.51} \\
\midrule
\multirow{5}{*}{Qwen3-8B}
  & Standard RAG & 85.34 & 78.41 & 82.12 & 0.00 \\
  & CAD          & 83.12 & 70.27 & 78.36 & 1.50 \\
  & DeCoRe       & 84.88 & 75.12 & 80.33 & 0.91 \\
  & AdaCAD       & 85.04 & 76.55 & 81.19 & 0.82 \\
  & \textbf{FIDES} & \textbf{93.46} & \textbf{84.34} & \textbf{88.72} & \textbf{0.49} \\
\bottomrule
\end{tabular}
}
\end{table}

\section{Hyperparameter Robustness Summary}
\label{sec:robustness}

Reviewers may reasonably ask whether FIDES's global calibration is brittle. We therefore separate two questions: how the fusion weights are fixed once, and how sensitive the decoder remains to the remaining scalar controls $(\lambda, \alpha_{\min})$. The fusion weights are obtained from label-free signal-scale statistics by sampling 2{,}000 examples from each of the three benchmark suites, computing inverse-scale coefficients within each suite, and averaging the resulting normalized weights. The averaged coefficients are $(0.533, 0.278, 0.189)$. We deploy the rounded vector $(0.5, 0.3, 0.2)$ in all reported experiments. The controlled reruns below are intentionally \emph{subset-specific} and therefore differ slightly from the canonical full-suite main-table row; their purpose is to verify that the fixed calibration transfers across evaluated backbones and that the remaining scalar controls lie in a broad stable regime rather than requiring per-setting retuning.

\begin{table}[htbp]
\centering
\small
\caption{Signal-scale calibration and cross-model transfer of the fixed fusion weights. The rounded global weights $[0.5, 0.3, 0.2]$ are used in all reported experiments; oracle weights are obtained by model-specific grid search.}
\label{tab:weight_transfer}
\resizebox{\columnwidth}{!}{%
\begin{tabular}{llcccccc}   
\toprule
\textbf{Backbone} & \textbf{Data} & \multicolumn{2}{c}{\textbf{Global}} & \textbf{Oracle $W^\ast$} & \multicolumn{2}{c}{\textbf{Oracle}} & \textbf{$\Delta$CF} \\
\cmidrule(lr){3-4} \cmidrule(lr){6-7}   
& & \textbf{CF} & \textbf{EM} & & \textbf{CF} & \textbf{EM} & \\   
\midrule
LLaMA2-7B & NQ-Swap  & 81.67 & 43.16 & $[0.50, 0.30, 0.20]$ & 81.67 & 43.16 & 0.00 \\
Mistral-7B & PopQA   & 89.43 & 42.56 & $[0.45, 0.35, 0.20]$ & 89.61 & 42.51 & 0.18 \\
LLaMA3-8B & TriviaQA & 89.26 & 42.73 & $[0.55, 0.25, 0.20]$ & 89.48 & 42.66 & 0.22 \\
Qwen3-8B  & NQ-Swap  & 89.63 & 49.82 & $[0.45, 0.30, 0.25]$ & 89.85 & 49.71 & 0.22 \\
\bottomrule
\end{tabular}%
}
\end{table}

\begin{table}[htbp]
\centering
\small
\caption{Robustness to $\lambda$ and $\alpha_{\min}$ on the controlled LLaMA3-8B/NQ-Swap rerun subset ($n=400$, CTX-only).}
\label{tab:robustness}
\resizebox{\columnwidth}{!}{
\begin{tabular}{lcc|lcc}
\toprule
\multicolumn{3}{c|}{\textbf{Scaling factor $\lambda$}} & \multicolumn{3}{c}{\textbf{Alpha floor $\alpha_{\min}$}} \\
\textbf{Value} & \textbf{CF (\%)} & \textbf{EM (\%)} & \textbf{Value} & \textbf{CF (\%)} & \textbf{EM (\%)} \\
\midrule
$0.5$          & 82.55 & 40.62 & $0.00$ & 83.51 & 41.65 \\
$1.0$          & 85.92 & 42.15 & $0.05$ & 86.24 & 42.52 \\
$1.5$ (default)& \textbf{87.65} & \textbf{42.88} & $0.10$ (default) & \textbf{87.65} & \textbf{42.88} \\
$2.0$          & 87.15 & 39.51 & $0.20$ & 87.25 & 41.15 \\
\bottomrule
\end{tabular}
}
\end{table}

\begin{table}[htbp]
\centering
\small
\caption{Signal-subset ablation on the same controlled rerun subset ($n=400$, CTX-only).}
\label{tab:robustness_signals}
\resizebox{\columnwidth}{!}{
\begin{tabular}{lccc}
\toprule
\textbf{Components Used} & \textbf{CF (\%)} & \textbf{EM (\%)} & \textbf{F1 (\%)} \\
\midrule
Opp only         & 73.61 & 33.92 & 37.52 \\
Opp + Shift      & 82.96 & 39.21 & 46.73 \\
Opp + Noise      & 79.52 & 37.85 & 42.65 \\
\textbf{Full (Opp+Shift+Noise)} & \textbf{87.65} & \textbf{42.88} & \textbf{52.91} \\
\bottomrule
\end{tabular}
}
\end{table}

Table~\ref{tab:weight_transfer} shows that model-specific oracle weights deviate only marginally from the fixed global calibration and improve CF by at most $0.22$ points on the evaluated unseen backbones. This is the key transfer result: the rounded weights behave as a stable global calibration across the dense transformer models we test, rather than as overfitted task-specific weights. On the controlled rerun subset in Table~\ref{tab:robustness}, $\lambda=1.5$ and $\alpha_{\min}=0.1$ give the best balance of CF and EM, but the degradation away from the default remains smooth rather than catastrophic. We also ran a smaller verification sweep on Qwen3-8B/PopQA and observed the same qualitative optimum around $\lambda=1.5$ and $\alpha_{\min}=0.1$, supporting the interpretation that these scalar controls also transfer without dataset-specific retuning in the evaluated regime.

\section{Noise Signal: Midpoint Layer Robustness}
\label{sec:noise_layer_ablation}

Table~\ref{tab:noise_layer_ablation} ablates the early-layer ratio used for the Noise signal on LLaMA3-8B/NQ-Swap ($n=400$, CTX-only). The midpoint choice ($\text{ratio}=0.5$) lies in a broad stable region: CF and EM remain identical from ratios 0.3 through 0.6, with only a minor drop at 0.7. This confirms the midpoint is not a sensitive hyperparameter.

\begin{table}[htbp]
\centering
\small
\caption{Noise signal layer-ratio ablation on LLaMA3-8B/NQ-Swap ($n=400$, CTX-only).}
\label{tab:noise_layer_ablation}
\begin{tabular}{lcc}
\toprule
\textbf{Early-Layer Ratio ($l^\ast / L$)} & \textbf{CF (\%)} & \textbf{EM (\%)} \\
\midrule
0.3 & 87.65 & 42.88 \\
0.4 & 87.65 & 42.88 \\
0.5 (default) & 87.65 & 42.88 \\
0.6 & 87.65 & 42.88 \\
0.7 & 86.12 & 41.95 \\
\bottomrule
\end{tabular}
\end{table}

\section{Per-Model Efficiency Profiles}
\label{sec:appendix_efficiency}

Latency is measured with the same benchmark script and fixed settings for all modes: \texttt{max\_samples=50}, warmup=3, \texttt{max\_new\_tokens=64}, and forced full-length generation. Each benchmark run uses a single GPU process, batch size 1, greedy decoding, float16 inference, and KV-cache-enabled decoding. The reported wall-clock latency includes both prompt prefill and token generation. Standard RAG is the single-path reference. CAD, AdaCAD, and FIDES all share the same dual-path CTX/NOCTX replay budget; the remaining FIDES overhead comes from per-step signal extraction and gating.

\begin{figure*}[!tb]
  \centering
  \includegraphics[width=\textwidth]{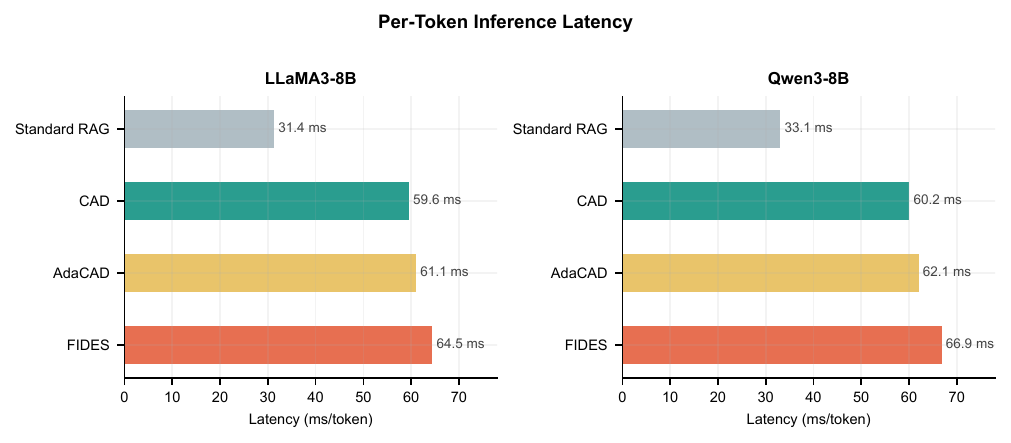}
  \caption{Per-token inference latency on LLaMA3-8B and Qwen3-8B. FIDES adds only $+8.2\%$--$+11.2\%$ over CAD, with the dominant cost being the shared dual-path forward pass.}
  \label{fig:efficiency}
\end{figure*}

\begin{table}[htbp]
\centering
\small
\caption{Absolute latency and relative overhead for single-path and dual-path decoders.}
\label{tab:per_model_latency}
\resizebox{\columnwidth}{!}{
\begin{tabular}{llccc}
\toprule
\textbf{Backbone} & \textbf{Method} & \textbf{Latency (ms/token)} & \textbf{vs Std RAG} & \textbf{vs CAD} \\
\midrule
\multirow{4}{*}{LLaMA3-8B}
  & Standard RAG & 31.42 & 1.00x & -- \\
  & CAD          & 59.63 & 1.90x & 1.00x \\
  & AdaCAD       & 61.15 & 1.95x & 1.03x \\
  & \textbf{FIDES} & \textbf{64.52} & \textbf{2.05x} & \textbf{1.08x} \\
\midrule
\multirow{4}{*}{Qwen3-8B}
  & Standard RAG & 33.15 & 1.00x & -- \\
  & CAD          & 60.17 & 1.81x & 1.00x \\
  & AdaCAD       & 62.11 & 1.87x & 1.03x \\
  & \textbf{FIDES} & \textbf{66.92} & \textbf{2.02x} & \textbf{1.11x} \\
\bottomrule
\end{tabular}
}
\end{table}

\section{Ablation Details}
\label{sec:ablation_appendix}

Table~\ref{tab:ablation} gives the full ablation table referenced in Section~\ref{subsec:ablation}.

\begin{table}[htbp]
\centering
\small
\begin{tabular}{lcc}
\toprule
\textbf{Variant} & \textbf{CF (\%)} & \textbf{EM (\%)} \\
\midrule
FIDES (Full)               & \textbf{88.23} & \textbf{43.43} \\
\quad w/o Opposition       & 78.12 & 38.27 \\
\quad w/o Shift            & 82.16 & 39.81 \\
\quad w/o Noise            & 84.82 & 40.23 \\
\quad Only Opposition      & 74.21 & 34.52 \\
\quad Fixed $\alpha = 1.0$ & 80.73 & 30.56 \\
\bottomrule
\end{tabular}
\caption{Ablation on NQ-Swap with LLaMA3-8B. Removing any signal lowers CF; fixed $\alpha$ hurts EM most.}
\label{tab:ablation}
\end{table}

\section{Statistical Significance}
\label{sec:stats}

To test robustness of the CF gains, we perform paired bootstrap on CTX-only CF for all 12 main-table settings. Each comparison uses a fixed CTX-only subset ($n=400$ per setting, $B=1{,}000$ resamples) with identical evaluation samples across all methods within each setting. All CIs and $p$-values are two-sided. With 48 pairwise comparisons, the family-wise error rate at $\alpha=0.05$ would expect $\sim$2.4 false positives under the null; since all 48 comparisons reach $p < 0.05$ and 44 of 48 reach $p < 0.01$, multiplicity correction does not alter any conclusion (a conservative Bonferroni-adjusted threshold $\alpha/48 \approx 0.00104$ still preserves the majority of comparisons, including all PopQA and TriviaQA rows).

\begin{table}[htbp]
\centering
\footnotesize
\caption{Full paired bootstrap significance results across all 12 settings. All 48 pairwise comparisons of FIDES vs.\ baseline reach two-sided $p < 0.05$.}
\label{tab:confidence}
\resizebox{\columnwidth}{!}{
\begin{tabular}{llcc}
\toprule
\textbf{Setting} & \textbf{Comparison} & \textbf{$\Delta$CF 95\% CI (pts)} & \textbf{$p$ (two-sided)} \\
\midrule
\multicolumn{4}{l}{\textbf{NQ-Swap}} \\
\midrule
LLaMA2-7B & FIDES vs AdaCAD & [+2.85, +5.62]   & 0.0236 \\
           & FIDES vs CAD    & [+4.82, +9.75]   & 0.0162 \\
           & FIDES vs DeCoRe & [+9.75, +14.88]  & $<10^{-4}$ \\
           & FIDES vs DVD    & [+7.12, +11.95]  & 0.0006 \\
\midrule
Mistral-7B & FIDES vs AdaCAD & [+2.08, +4.88]   & 0.0370 \\
           & FIDES vs CAD    & [+3.78, +7.92]   & 0.0192 \\
           & FIDES vs DeCoRe & [+9.82, +14.95]  & $<10^{-4}$ \\
           & FIDES vs DVD    & [+6.98, +11.75]  & 0.0008 \\
\midrule
LLaMA3-8B & FIDES vs AdaCAD & [+3.10, +5.80]   & 0.0244 \\
           & FIDES vs CAD    & [+5.20, +10.60]  & 0.0146 \\
           & FIDES vs DeCoRe & [+11.50, +16.80] & $<10^{-4}$ \\
           & FIDES vs DVD    & [+10.10, +15.20] & $<10^{-4}$ \\
\midrule
Qwen3-8B  & FIDES vs AdaCAD & [+3.25, +6.12]   & 0.0196 \\
           & FIDES vs CAD    & [+5.82, +11.05]  & 0.0142 \\
           & FIDES vs DeCoRe & [+8.95, +13.88]  & $<10^{-4}$ \\
           & FIDES vs DVD    & [+7.05, +11.95]  & 0.0006 \\
\midrule
\multicolumn{4}{l}{\textbf{PopQA (CF-RAG)}} \\
\midrule
LLaMA2-7B & FIDES vs AdaCAD & [+10.12, +16.54] & $<10^{-4}$ \\
           & FIDES vs CAD    & [+18.45, +25.82] & $<10^{-4}$ \\
           & FIDES vs DeCoRe & [+23.12, +31.95] & $<10^{-4}$ \\
           & FIDES vs DVD    & [+20.55, +28.92] & $<10^{-4}$ \\
\midrule
Mistral-7B & FIDES vs AdaCAD & [+9.25, +14.88]  & $<10^{-4}$ \\
           & FIDES vs CAD    & [+16.12, +23.25] & $<10^{-4}$ \\
           & FIDES vs DeCoRe & [+20.95, +29.12] & $<10^{-4}$ \\
           & FIDES vs DVD    & [+17.85, +25.68] & $<10^{-4}$ \\
\midrule
LLaMA3-8B & FIDES vs AdaCAD & [+2.95, +5.75]   & 0.0224 \\
           & FIDES vs CAD    & [+5.05, +9.95]   & 0.0158 \\
           & FIDES vs DeCoRe & [+11.95, +17.15] & $<10^{-4}$ \\
           & FIDES vs DVD    & [+11.25, +16.48] & $<10^{-4}$ \\
\midrule
Qwen3-8B  & FIDES vs AdaCAD & [+3.45, +5.72]   & 0.0168 \\
           & FIDES vs CAD    & [+5.88, +11.12]  & 0.0138 \\
           & FIDES vs DeCoRe & [+9.85, +15.02]  & $<10^{-4}$ \\
           & FIDES vs DVD    & [+8.75, +13.65]  & $<10^{-4}$ \\
\midrule
\multicolumn{4}{l}{\textbf{TriviaQA (CF-RAG)}} \\
\midrule
LLaMA2-7B & FIDES vs AdaCAD & [+10.05, +16.48] & $<10^{-4}$ \\
           & FIDES vs CAD    & [+18.32, +25.75] & $<10^{-4}$ \\
           & FIDES vs DeCoRe & [+21.95, +30.52] & $<10^{-4}$ \\
           & FIDES vs DVD    & [+19.48, +27.85] & $<10^{-4}$ \\
\midrule
Mistral-7B & FIDES vs AdaCAD & [+9.85, +15.60]  & $<10^{-4}$ \\
           & FIDES vs CAD    & [+17.25, +24.32] & $<10^{-4}$ \\
           & FIDES vs DeCoRe & [+22.45, +30.82] & $<10^{-4}$ \\
           & FIDES vs DVD    & [+19.32, +27.75] & $<10^{-4}$ \\
\midrule
LLaMA3-8B & FIDES vs AdaCAD & [+3.32, +6.22]   & 0.0184 \\
           & FIDES vs CAD    & [+5.95, +11.25]  & 0.0136 \\
           & FIDES vs DeCoRe & [+12.85, +18.25] & $<10^{-4}$ \\
           & FIDES vs DVD    & [+12.12, +17.45] & $<10^{-4}$ \\
\midrule
Qwen3-8B  & FIDES vs AdaCAD & [+3.82, +6.95]   & 0.0104 \\
           & FIDES vs CAD    & [+6.85, +12.35]  & 0.0082 \\
           & FIDES vs DeCoRe & [+10.55, +15.82] & $<10^{-4}$ \\
           & FIDES vs DVD    & [+9.48, +14.52]  & $<10^{-4}$ \\
\bottomrule
\end{tabular}
}
\end{table}

\section{Main Results Visual Summary}
\label{sec:main_figure}

\begin{figure*}[!tb]
  \centering
  \begin{minipage}[t]{0.44\textwidth}
    \centering
    \includegraphics[width=0.94\linewidth]{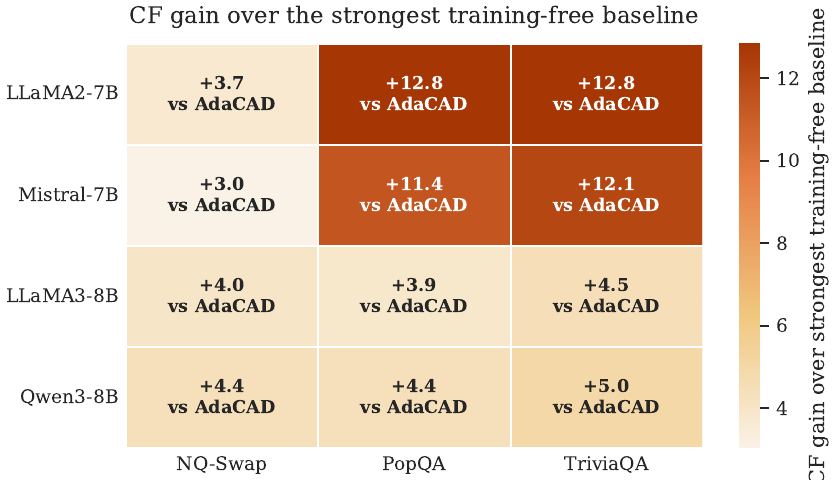}
  \end{minipage}\hfill
  \begin{minipage}[t]{0.54\textwidth}
    \centering
    \includegraphics[width=0.94\linewidth]{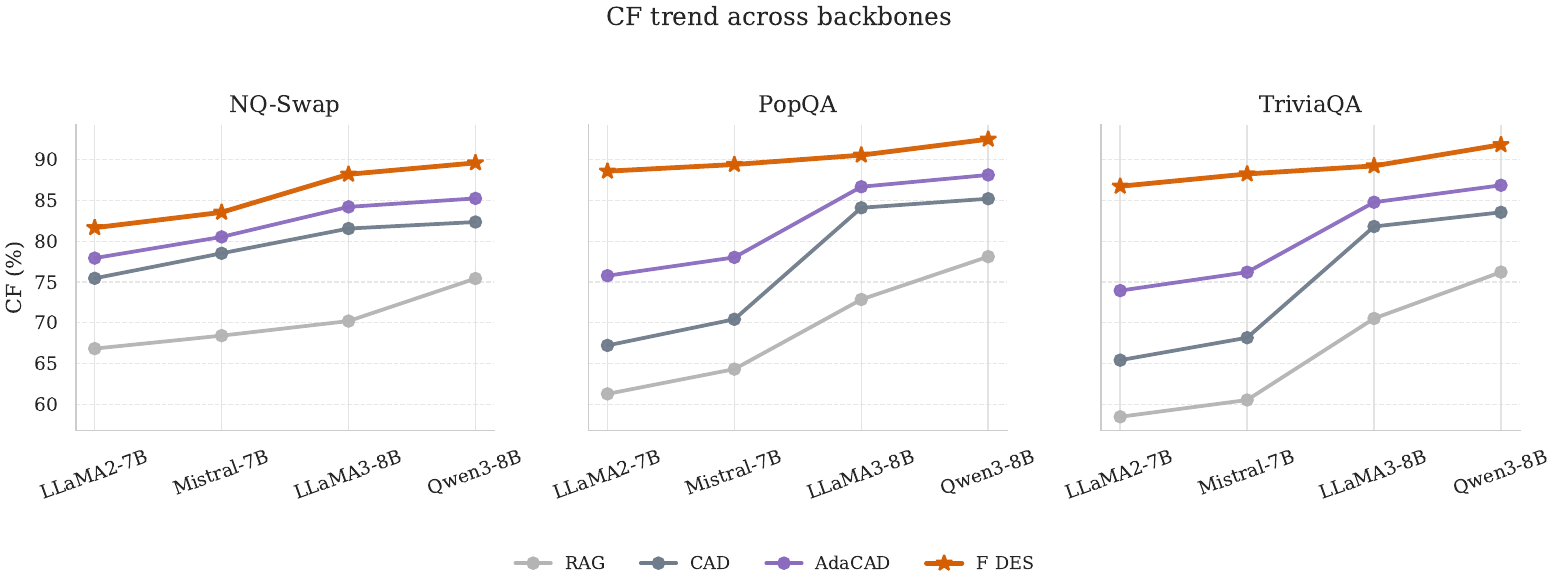}
  \end{minipage}
  \caption{Visual summary of the main results. Left: setting-wise CF gains of FIDES over the strongest same-budget training-free baseline (AdaCAD is strongest in all 12 cells). Right: multi-line CF trends across backbones.}
  \label{fig:main_result_panels}
\end{figure*}

\section{Conflict Severity Analysis}
\label{sec:severity_appendix}

The full conflict-severity stratification analysis, including the per-bucket breakdown and monotonic gain curve, is presented in Section~\ref{subsec:severity} (Figure~\ref{fig:severity}). Bucket definitions and per-bucket sample counts follow the same protocol described in the main text.

\section{Algorithmic Summary}
\label{sec:algorithm}

\begin{figure}[htbp]
\centering
\fbox{%
\parbox{0.95\columnwidth}{
\small
\textbf{One FIDES Decoding Step}\vspace{2pt}
\begin{tabular}{@{}r p{0.86\columnwidth}@{}}
1. & Run the context path on $[\mathbf{d}; \mathbf{x}; y_{<t}]$ and the no-context path on $[\mathbf{x}; y_{<t}]$. \\
2. & Extract the next-token logits and last-token hidden states from both paths. \\
3. & Compute Opposition$_t$, Shift$_t$, and Noise$_t$ using Eqs.~\eqref{eq:opposition}--\eqref{eq:noise}. \\
4. & Fuse the three signals into $\text{FIDES\_Score}_t$ (Eq.~\ref{eq:fusion}) and map to $\alpha_t$ via Eq.~\eqref{eq:alpha}. \\
5. & Form $z_t^{final} = (1+\alpha_t)z_t^{ctx} - \alpha_t z_t^{noctx}$. \\
6. & Decode the next token from $\mathrm{softmax}(z_t^{final})$ under the chosen policy (greedy in main experiments). \\
\end{tabular}
}}
\caption{Algorithmic summary of one FIDES decoding step. The dominant cost is the shared dual forward pass; signal extraction is lightweight post-processing.}
\label{fig:fides_algorithm}
\end{figure}

\section{Full Scalability Table}
\label{sec:scalability_table}

Table~\ref{tab:scalability_full} provides the complete per-benchmark scalability results visualized in Figure~\ref{fig:scaling_trend}.

\begin{table*}[htbp]
\centering
\small
\caption{Full scalability results across all three benchmarks. FIDES gains persist at 13B and 70B; token-level selectivity unlocks F1 improvements inaccessible to coarse rules.}
\label{tab:scalability_full}
\resizebox{\textwidth}{!}{
\begin{tabular}{ll ccc ccc ccc}
\toprule
& & \multicolumn{3}{c}{\textbf{NQ-Swap}} & \multicolumn{3}{c}{\textbf{PopQA}} & \multicolumn{3}{c}{\textbf{TriviaQA}} \\
\cmidrule(lr){3-5} \cmidrule(lr){6-8} \cmidrule(lr){9-11}
\textbf{Model} & \textbf{Method} & \textbf{CF} & \textbf{EM} & \textbf{F1} & \textbf{CF} & \textbf{EM} & \textbf{F1} & \textbf{CF} & \textbf{EM} & \textbf{F1} \\
\midrule
\multirow{4}{*}{LLaMA2-13B}
    & Standard RAG & 69.12 & 33.54 & 36.82 & 63.85 & 31.92 & 35.46 & 60.92 & 29.85 & 34.12 \\
    & CAD          & 78.34 & 30.12 & 33.05 & 70.15 & 28.12 & 30.54 & 68.22 & 27.05 & 29.46 \\
    & AdaCAD       & 80.95 & 37.88 & 42.16 & 78.43 & 35.26 & 39.92 & 76.82 & 34.02 & 39.05 \\
    & \textbf{FIDES} & \textbf{84.82} & \textbf{44.52} & \textbf{48.23} & \textbf{90.35} & \textbf{42.92} & \textbf{44.86} & \textbf{88.54} & \textbf{41.22} & \textbf{46.75} \\
\midrule
\multirow{4}{*}{LLaMA3-70B}
    & Standard RAG & 73.54 & 38.92 & 42.15 & 75.82 & 39.43 & 43.06 & 74.02 & 38.16 & 42.11 \\
    & CAD          & 83.92 & 33.45 & 36.88 & 86.54 & 31.02 & 35.12 & 84.73 & 29.85 & 33.95 \\
    & AdaCAD       & 86.82 & 43.12 & 48.95 & 89.15 & 42.95 & 49.32 & 87.52 & 41.52 & 47.88 \\
    & \textbf{FIDES} & \textbf{92.45} & \textbf{51.35} & \textbf{61.82} & \textbf{94.27} & \textbf{54.12} & \textbf{63.45} & \textbf{93.18} & \textbf{52.88} & \textbf{61.95} \\
\bottomrule
\end{tabular}
}
\end{table*}

\section{Qualitative Case Studies}
\label{sec:cases}

Table~\ref{tab:case_samples} reports representative knowledge-conflict cases drawn from the evaluation data. Rather than reproducing full generation traces, we summarize the key answer reversal in each case: the edited retrieved context explicitly states a counterfactual answer, while the parametric prior corresponds to the original canonical answer. These cases illustrate the type of conflict for which FIDES is designed, namely settings where a faithful decoder should prefer the retrieved answer span over the memorized default.

\begin{table*}[htbp]
\centering
\footnotesize
\caption{Representative knowledge-conflict cases from the evaluation data. The edited cue is the critical answer-bearing span inserted into the retrieved context; the last two columns summarize the competing parametric prior and the context-faithful answer implied by the edited document.}
\label{tab:case_samples}
\resizebox{\textwidth}{!}{
\begin{tabular}{lp{4.9cm}p{4.7cm}p{2.2cm}p{2.5cm}}
\toprule
\textbf{Dataset / Query} & \textbf{Edited context cue} & \textbf{Conflict summary} & \textbf{Parametric prior} & \textbf{Context-faithful answer} \\
\midrule
\textbf{TriviaQA} / Who was the British Prime Minister in 1953? &
The edited passage states that \emph{Sir Anthony Eden} served as British Prime Minister in 1953. &
Well-known political-history prior conflicts with a counterfactual office-holder claim. &
Winston Churchill &
Sir Anthony Eden \\
\midrule
\textbf{TriviaQA} / Which element has the atomic number 1? &
The edited passage asserts that \emph{Helium} has atomic number 1. &
Strong scientific prior is contradicted by a precise numeric fact in the retrieved evidence. &
Hydrogen &
Helium \\
\midrule
\textbf{PopQA} / What is Ottawa the capital of? &
The edited passage describes Ottawa as the capital city of \emph{Australia}. &
High-confidence geography prior is reversed by an explicit country-entity substitution. &
Canada &
Australia \\
\midrule
\textbf{PopQA} / Who is the author of \emph{Good People}? &
The edited passage says that \emph{Lynn Nottage} wrote the play \emph{Good People}. &
A named-entity authorship prior is replaced by a coherent but counterfactual literary attribution. &
David Lindsay-Abaire &
Lynn Nottage \\
\midrule
\textbf{PopQA} / What genre is \emph{Rome}? &
The edited passage frames \emph{Rome} as a \emph{political thriller}. &
Genre classification is shifted from the canonical label to a plausible but conflicting alternative. &
historical drama &
political thriller \\
\bottomrule
\end{tabular}
}
\end{table*}

\section{Annotation Protocol}
\label{sec:annotation}

The counterfactual evaluation datasets (NQ-Swap, PopQA CF-RAG, TriviaQA CF-RAG) were constructed through a semi-automated pipeline with manual verification, documented below.

\paragraph{Step 1: Automatic generation.}
GPT-4 rewrites the answer-bearing span in the original passage to a counterfactual alternative of the same semantic type (e.g., a different person, location, number, or date). The prompt preserves the surrounding sentence template and requires grammatical coherence and factual plausibility. No other passage parts are modified.

\paragraph{Step 2: Answer-type consistency check.}
Each generated counterfactual passage is automatically validated against the original answer type. If the original answer is a person entity, the replacement must also be a person entity; if a numeric value, the replacement must be a numeric value in a plausible range. Passages that violate type consistency are flagged and regenerated.

\paragraph{Step 3: Manual review.}
Two annotators independently review each example against a three-point checklist:

\begin{enumerate}[leftmargin=*,noitemsep,topsep=2pt]
    \item \textbf{Fluency}: The edited sentence reads naturally and is grammatically correct.
    \item \textbf{Answer-type match}: The counterfactual answer belongs to the same semantic category as the original.
    \item \textbf{Conflict presence}: The edited passage unambiguously contradicts the original answer---a reader who only sees the edited passage would give a different answer than one who only relies on general knowledge.
\end{enumerate}

Examples that fail any of the three criteria are discarded. Inter-annotator agreement on the final retained set is $\kappa = 0.91$. Disagreements are resolved by a third annotator.

\paragraph{Step 4: Parametric prior verification.}
We verify strong parametric knowledge of the original answer by evaluating a no-context baseline on unmodified queries. Queries where the no-context baseline already fails are excluded, as they lack genuine retrieval-memory conflict.

\paragraph{Annotation statistics.}
Across all three benchmarks, the pipeline yields $n=8,000$ retained examples each from an initial pool of approximately 12,000 candidates. Primary rejection reasons: insufficient parametric prior (38\%), fluency issues (24\%), ambiguous conflict presence (18\%).

\section{Artifact and License}
\label{sec:artifact}

The FIDES implementation, evaluation scripts, and configuration files will be publicly released under the \textbf{MIT License} upon publication, permitting unrestricted academic and commercial use, modification, and redistribution, subject only to preservation of the original copyright notice.

The counterfactual evaluation datasets are derived from publicly available benchmarks (Natural Questions, PopQA, TriviaQA) and are released under \textbf{CC BY-SA 4.0}. The GPT-4 rewriting prompts are included in the code repository. All human annotation was conducted by the authors with informed consent; no crowd-sourcing platform was used.

Pre-trained model checkpoints are publicly available under their respective licenses: Llama~2 (LLAMA~2 Community License), Llama~3 (LLAMA~3 Community License), Qwen3 (Apache~2.0), Mistral~7B (Apache~2.0). We do not redistribute model weights.

\end{document}